%% file: conll-2019.tex
\documentclass[11pt,a4paper]{article}
\usepackage[hyperref]{conll-2019}
\usepackage{times}
\usepackage{latexsym}

\usepackage{url}
\usepackage{graphicx}
\usepackage{amsmath}
\usepackage{enumitem}
\usepackage{booktabs}
\usepackage{relsize}
\usepackage{stfloats}

\aclfinalcopy 

\newcommand{\ourmodel}{G-DuHA}

\title{Goal-Embedded Dual Hierarchical Model \\for Task-Oriented Dialogue Generation}

\author{Yi-An Lai \\
  Amazon ML \\
  Seattle \\
  {\tt yianl@amazon.com} \\\And
  Arshit Gupta \\
  Amazon ML \\
  Seattle \\
  {\tt arshig@amazon.com} \\\And
  Yi Zhang \\
  Amazon ML \\
  Seattle \\
  {\tt yizhngn@amazon.com} \\}

\date{}

\begin{document}
\maketitle
\begin{abstract}
  Hierarchical neural networks are often used to model inherent structures within dialogues.
  For goal-oriented dialogues, these models miss a mechanism adhering to the goals and neglect the distinct conversational patterns between two interlocutors.
  In this work, we propose {\it Goal-Embedded Dual Hierarchical Attentional Encoder-Decoder (\ourmodel)} able to center around goals and capture interlocutor-level disparity while modeling goal-oriented dialogues.
  Experiments on dialogue generation, response generation, and human evaluations demonstrate that the proposed model successfully generates higher-quality, more diverse and goal-centric dialogues. Moreover, we apply data augmentation via goal-oriented dialogue generation for task-oriented dialog systems with better performance achieved.
  
\end{abstract}

\section{Introduction}
Modeling a probability distribution over word sequences is a core topic in natural language processing, with language modeling being a flagship problem and mostly tackled via recurrent neural networks (RNNs) \cite{mikolov2012context, melis2017state, Merity2018RegularizingAO}.

Recently, dialogue modeling has drawn much attention with applications to response generation \cite{Serban2016MultiresolutionRN, Li2016DeepRL, asghar2018affective} or data augmentation \cite{Yoo2019DataAF}.
It's inherently different from language modeling as the conversation is conducted in a turn-by-turn nature.
\cite{Serban2016BuildingED} imposes a hierarchical structure on encoder-decoder to model this utterance-level and dialogue-level structures, followed by \cite{Serban2016AHL, chen2018hierarchical, le2018variational}.

\begin{figure}[t!]
  \centering
    \includegraphics[width=0.48\textwidth]{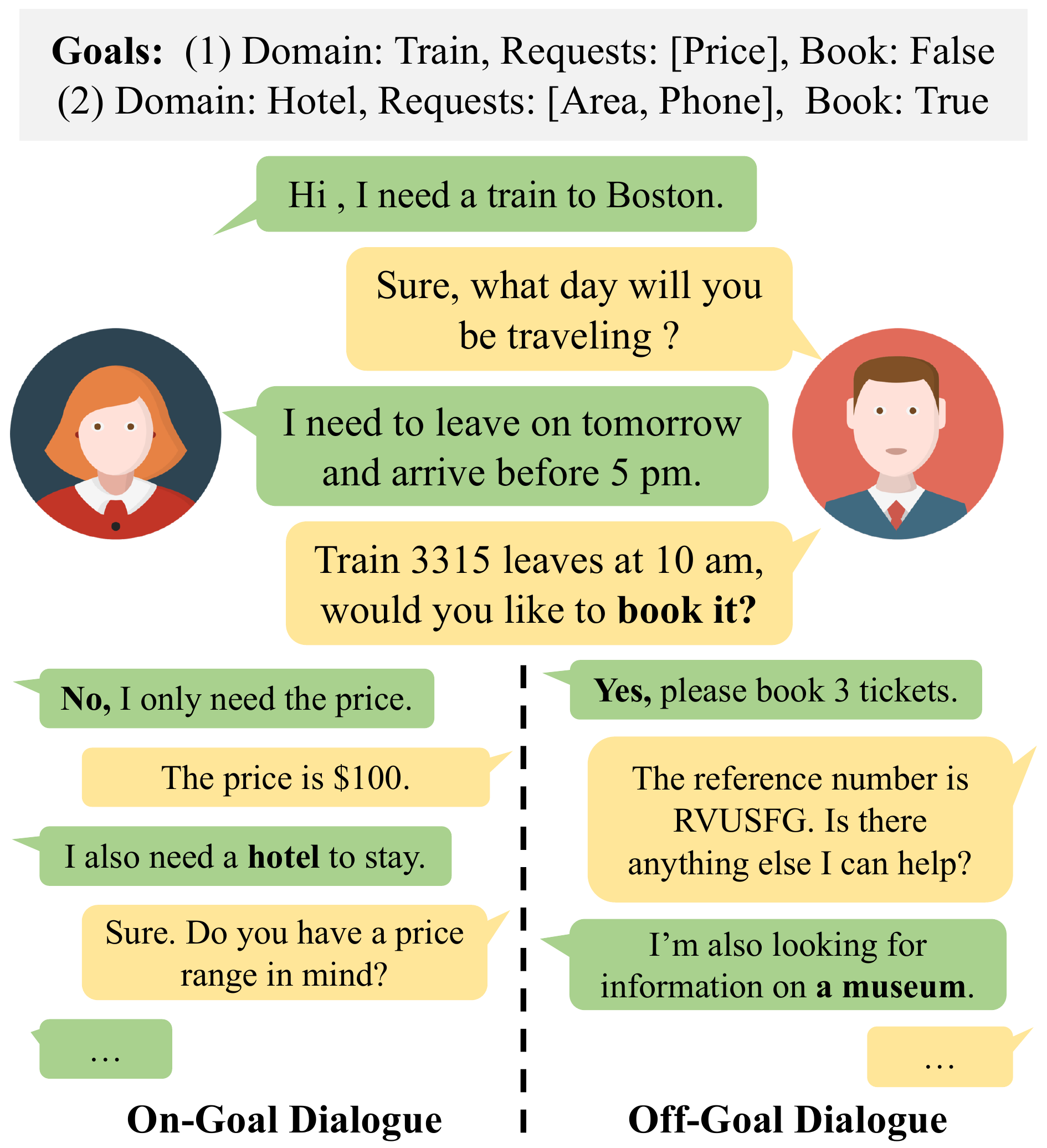}
  \caption{On-goal dialogues follow the given goals such as no booking of train tickets and a hotel reservation. Off-goal dialogues have context switches to other domains non-relevant to goals.}
  \label{fig:motivation}
\end{figure}

However, when modeling dialogues involving two interlocutors center around one or more goals, these systems generate utterances with the greatest likelihood but without a mechanism sticking to the goals.
This makes them go off the rails and fail to model context-switching of goals.
Most of the generated conversations become off-goal dialogues with utterances being non-relevant or contradicted to goals rather than on-goal dialogues. The differences are illustrated in Figure \ref{fig:motivation}.

Besides, two interlocutors in a goal-oriented dialogue often play distinct roles as one has requests or goals to achieve and the other provides necessary support.
Modeled by a single hierarchical RNN, this interlocutor-level disparity is neglected and constant context switching of roles could reduce the capacity for tracking conversational flow and long-term temporal structure.

To resolve the aforementioned issues when modeling goal-oriented dialogues, we propose the Goal-Embedded Dual Hierarchical Attentional Encoder-Decoder (\ourmodel) to tackle the problems via three key features.
First, the goal embedding module summarizes one or more goals of the current dialogue as goal contexts for the model to focus on across a conversation.
Second, the dual hierarchical encoder-decoders can naturally capture interlocutor-level disparity and represent interactions of two interlocutors.
Finally, attentions are introduced on word and dialogue levels to learn temporal dependencies more easily.

In this work, our contributions are that we propose a model called goal-embedded dual hierarchical attentional encoder-decoder (\ourmodel) to be the first model able to focus on goals and capture interlocutor-level disparity while modeling goal-oriented dialogues. 
With experiments on dialogue generation, response generation and human evaluations, we demonstrate that our model can generate higher-quality, more diverse and goal-focused dialogues.
In addition, we leverage goal-oriented dialogue generation as data augmentation for task-oriented dialogue systems, with better performance achieved.

\section{Related Work}

Dialogues are sequences of utterances, which are sequences of words. 
For modeling or generating dialogues, hierarchical architectures are usually used to capture their conversational nature.
Traditionally, language models are also used for modeling and generating word sequences.
As goal-oriented dialogues are generated, they can be used in data augmentation for task-oriented dialogue systems.
We review related works in these fields.

\vspace{1.5mm}
\noindent
{\bf Dialogue Modeling.}
To model conversational context and turn-by-turn structure of dialogues, \cite{Serban2016BuildingED} devised hierarchical recurrent encoder-decoder (HRED).
Reinforcement and adversarial learning are then adopted to improve naturalness and diversity \cite{Li2016DeepRL, Li2017AdversarialLF}.
Integrating HRED with the latent variable models such as variational autoencoder (VAE) \cite{Kingma2014AutoEncodingVB} extends another line of advancements \cite{Serban2016AHL, Zhao2017LearningDD, Park2018AHL, Le2018VariationalME}.
However, these systems are not designed for task-oriented dialogue modeling as goal information is not considered.
Besides, conversations between two interlocutors are captured with a single encoder-decoder by these systems.

\vspace{1.5mm}
\noindent
{\bf Language Modeling.} A probability distribution of a word sequence $w_{1:T}=(w_1, w_2, ..., w_T)$ can be factorized as $p(w_1) \prod_{t=2}^{T} p(w_t|w_{1:t-1})$.
To approximate the conditional probability $p(w_t|w_{1:t-1})$, counted statistics and smoothed N-gram models have been used before \cite{goodman2001bit, katz1987estimation, kneser1995improved}.
Recently, RNN-based models have achieved a better performance \cite{mikolov2010recurrent, Jzefowicz2016ExploringTL, Grave2017ImprovingNL, Melis2018OnTS}.
As conversational nature is not explicitly modeled, models often have role-switching issues.

\vspace{1.5mm}
\noindent
{\bf Task-Oriented Dialogue Systems.} Conventional task-oriented dialog systems entails a sophisticated pipeline \cite{raux2005let, Young2013POMDPBasedSS} with components including spoken language understanding \cite{chen2016end, Mesnil2015UsingRN, gupta2019simple}, dialog state tracking \cite{henderson2014word, Mrksic2017NeuralBT}, and dialog policy learning \cite{su2016line, gavsic2014gaussian}.
Building a task-oriented dialogue agent via end-to-end approaches has been explored recently \cite{Li2017EndtoEndTN, Wen2017ANE}.
Although several conversational datasets are published recently \cite{Gopalakrishnan2019, Henderson2019},
the scarcity of annotated conversational data remains a key problem when developing a dialog system.
This motivates us to model task-oriented dialogues with goal information in order to achieve controlled dialogue generation for data augmentation.

\section{Model Architecture}

\begin{figure*}
  \centering
    \includegraphics[width=0.8\textwidth]{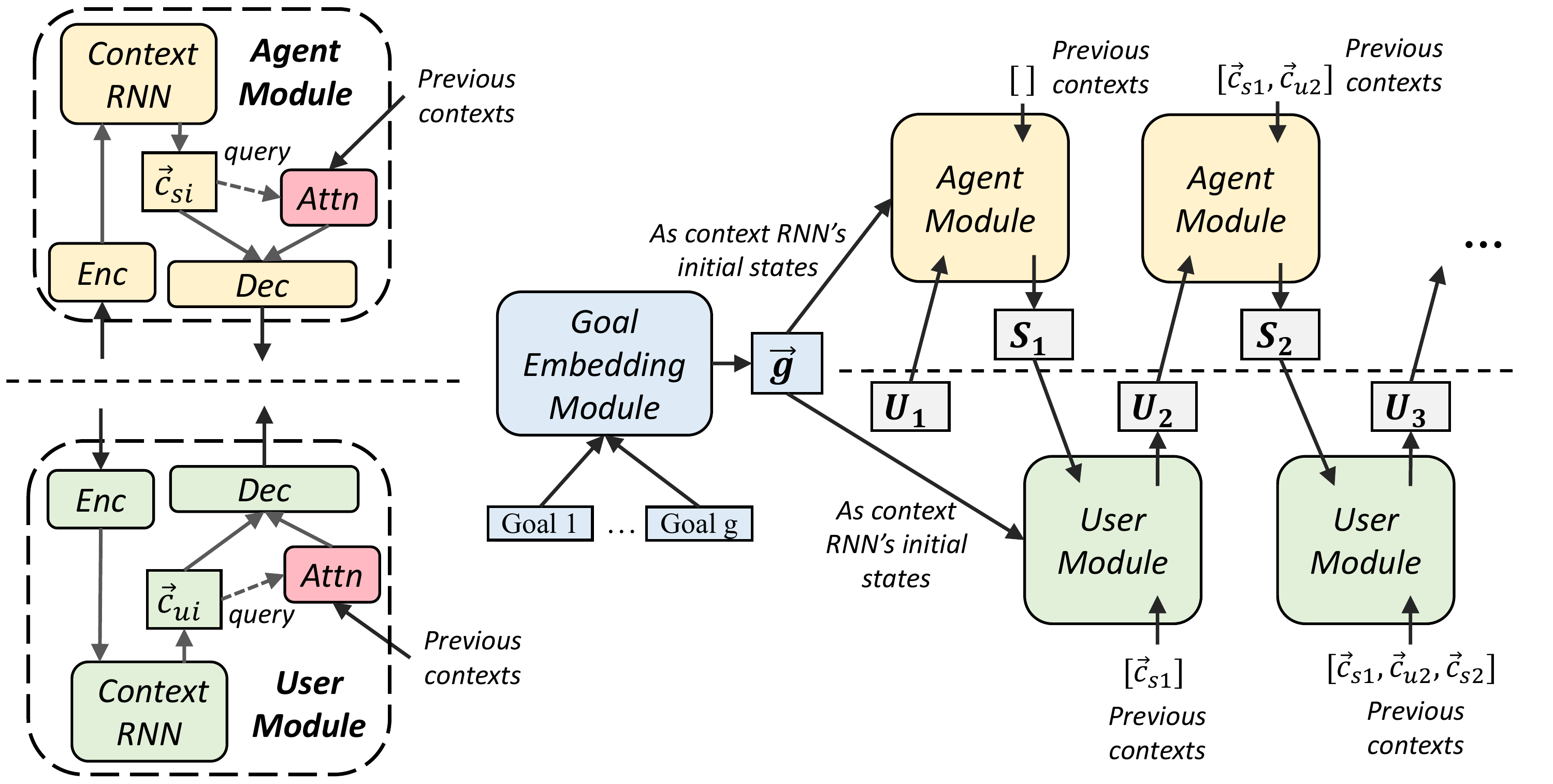}
  \caption{{\ourmodel} architecture. The goal embedding module embeds goals as priors for context RNNs. Dual hierarchical RNNs naturally model two interlocutors. An attention over previous contexts captures long-term dependencies. For encoders, word attentions are used to summarize local importance of words. (Enc: Encoder, Dec: Decoder, Attn: Attention, $U_i$: User's utterance, $S_i$: Agent's utterance)}
  \label{fig:model}
\end{figure*}

Given a set of goals and the seed user utterance, we want to generate a goal-centric or on-goal dialogue that follows the domain contexts and corresponding requests specified in goals. 
In this section, we start with the mathematical formulation, then introduce our proposed model, and describe our model's training objective and inference.

At training time, $K$ dialogues $\{D_1, ..., D_K\}$ are given where each $D_i$ associates with $N_i$ goals ${\bf g_i} = \{g_{i1}, g_{i2}, ..., g_{iN_i}\}$. 
A dialogue $D_i$ consists of $M$ turns of utterances between a user ${\bf u}$ and a system agent ${\bf s}$ $({\bf w_{u1}}, {\bf w_{s1}}, {\bf w_{u2}}, {\bf w_{s2}}, ...)$, where ${\bf w_{u1}}$ is a word sequence $w_{u1, 1}, w_{u1, 2}, ..., w_{u1, N_{u1}}$ denoting the user's first utterance.

The task-oriented dialogue modeling aims to approximate the conditional probability of user's or agent's next utterance given previous turns and goals. 
It can be further decomposed over generated words, e.g. 
\begin{equation}
\begin{split}
P({\bf w_{um}}|{\bf w_{u1}}, {\bf w_{s1}}, ..., {\bf w_{s(m-1)}}, {\bf g_i}) = \\
    \quad \prod_{n=1}^{N_{um}} P(w_{um, n}|w_{um, <n}, {\bf w_{u1}}, ...)
\end{split}
\end{equation}

To model goal-oriented dialogues between two interlocutors, we propose Goal-embedded Dual Hierarchical Attentional Encoder-Decoder (\ourmodel) as illustrated in Fig. \ref{fig:model}.
Our model comprises goal embedding module, dual hierarchical RNNs, and attention mechanisms detailed below.


\subsection{Goal Embedding Module}
We represent each goal in $\{g_{i1}, g_{i2}, ...\}$ using a simple and straightforward binary or multi-one-hot encoding followed by a feed-forward network (FFN).
A goal could have a specific domain such as hotel or a request such as price or area, Table \ref{table:dialsample} shows a few examples.
Our goal embedding module, a FFN, then converts binary encoding of each goal into a goal embedding, where the FFN is learned during training.
If multiple goals present, all goal embeddings are then added up element-wisely to be the final goal embedding:

\vspace{-1mm}
\begin{equation}
\overrightarrow{g_i} = \sum_{j=1}^{|g_i|} FFN( Encode(g_{ij}) )
\end{equation}

The output of the goal embedding module has the same number of dimensions as context RNN's hidden state and is used as initial states for all layers of all context RNNs to inform the model about a set of goals to focus on.

\subsection{Dual Hierarchical Architecture}
The hierarchical encoder-decoder structure \cite{Serban2016BuildingED} is designed for utterance level and context level modeling.
With a single encoder, context RNN, and decoder, the same module is used to process input utterances, track contexts, and generate responses for both interlocutors.

In task-oriented dialogues, however, roles are distinct as the user aims to request information or make reservations to achieve goals in mind and the system agent provides necessary help.
To model this interlocutor-level disparity, we extend it into a dual architecture involving two hierarchical RNNs, each serves as a role in a dialogue.

\subsection{Attention Mechanisms}
At the utterance level, as importance of words can be context dependent, our model uses first hidden layer's states of the context RNN as the query for attention mechanisms \cite{Bahdanau2015NeuralMT, Xu2015ShowAA} to build an utterance representation. A feed-forward network is involved to computed attention scores, whose input is the concatenation of the query and an encoder output.

At the dialogue level, for faster training, our model has a skip connection to add up context RNN's raw output with its input as the final output $c_t$.
To model long-term dependencies, an attention module is applied to summarize all previous contexts into a global context vector. 
Specifically, a feed-forward network takes the current context RNN's output $c_t$ as the query and all previous context outputs from both context RNNs as keys and values to compute attention scores. 
The global context vector is then concatenated with $c_t$ to form our final context vector for decoder to consume.

\subsection{Objective}
For predicting the end of a dialogue, we exploit a feed-forward network over the final context vector for a binary prediction. 
Thus our training objective can be written as
\begin{equation}
L = \sum_{i = 1}^{K} \Big[ -\log p_{\boldsymbol{\theta}}(D_{i}) + \sum_{t}^{M_i} - \log p_{\boldsymbol{\theta}}^{end}(e_{it}) \Big],
\end{equation}
where our model $p_{\boldsymbol{\theta}}$ has parameters $\boldsymbol{\theta}$, $M_i$ is the number of turns, $e_{it}$ is $0$ as the dialogue $D_i$ continues and $1$ if it terminates at turn $t$.

\subsection{Generation}
At dialogue generation time, a set of goals $\{g_{i1}, g_{i2}, ...\}$ and a user utterance ${\bf w_{u1}}$ as a seed are given.
Then our model will generate conversations simulating interactions between a user and an agent that seek to complete all given goals.
The generation process terminates as the end of dialogue prediction outputs a positive or the maximum number of turns is reached.

\section{Experiments}

We evaluate our approach on dialogue generation and response generation as well as by humans.
Ablation studies and an extrinsic evaluation that leverages dialogue generation as a data augmentation method are reported  in the subsequent section.

\subsection{Dataset}
Experiments are conducted on a task-oriented human-human written conversation dataset called MultiWOZ \cite{Budzianowski2018MultiWOZA}, the largest publicly available dataset in the field.
Dialogues in the dataset span over diverse topics, one to more goals, and multiple domains such as \texttt{restaurant}, \texttt{hotel}, \texttt{train}, etc.
It consists of $8423$ train dialogues, $1000$ validation and $1000$ test dialogues with on average $15$ turns per dialogue and $14$ tokens per turn.

\subsection{Baselines}
We compare our approach against four baselines:

(i) LM+G: As long-established methods for language generation, we adopt an RNN language model (LM) 
with 3-layer 200-hidden-unit GRU \cite{Cho2014LearningPR} incorporating our goal embedding module as a baseline, which has goal information but no explicit architecture for dialogues.

(ii) LM+G-XL: To show the possible impact of model size, a larger LM that has a 3-layer 450-hidden-unit GRU is adopted as another baseline.

(iii) Hierarchical recurrent encoder-decoder (HRED)  \cite{Serban2016BuildingED}: As the prominent model for dialogues, we use HRED as the baseline that has a dialogue-specific architecture but no goal information.
The encoder, decoder, and context RNN are 2-layer 200-hidden-unit GRUs.

(iv) HRED-XL: We also use a larger HRED with 350 hidden units for all GRUs as a baseline to show the impact of model size.

\subsection{Implementation Details}
In all experiments, we adopt the delexicalized form of dialogues as shown in Table \ref{table:dialsample} with vocabulary size, including slots and special tokens, to be $4258$.
The max number of turns and sequence length are capped to $22$ and $36$, respectively.

{\ourmodel} uses 2-layer, 200-hidden-unit GRUs as all encoders, decoders, and context RNNs.
All feed-forward networks have 2 layers with non-linearity.
FFNs of encoder attention and end of dialogue prediction have $50$ hidden units.
The FFNs of context attention and goal embedding gets $100$ and $200$ hidden units.
We simply use greedy decoding for utterance generation.

All models initialize embeddings with pre-trained fast-text vectors on wiki-news \shortcite{mikolov2018advances} and are trained by the Adam optimizer \shortcite{Kingma2015AdamAM} with early-stopping to prevent overfitting.
To mitigate the discrepancy between training and inference, we pick predicted or ground-truth utterance as the current input uniformly at random when training.

\begin{table*}[t!]
\centering
\smaller[1]
\begin{tabular}{l|c|c c c c | c c c | c c c | c c }
  \hline
  {\bf Model} & {\bf Size} & {\bf BLEU} & {\bf B1}  & {\bf B2}  & {\bf B3}  & {\bf D-1}  & {\bf D-2}  & {\bf D-U} & {\bf P} & {\bf R} & {\bf F1} & {\bf L-D} & {\bf L-U}\\
  \hline
  \hline
  LM+G & 4.2 M    & 6.34 & 23.24 & 12.89 & 8.76 & 0.16 & 0.88 & 23.75 & 89.52 & 82.39 & {\bf 84.89} & 15.1 & 13.0\\
  LM+G-XL&8.2 M& 6.22 & 23.10 & 12.75 & 8.63 & 0.16 & 0.93 & {\bf 26.38} & {\bf 90.03} & 81.62 & 84.71 & 14.7 & 14.4\\
  HRED & 5.1 M       & 5.40 & 21.91 & 11.58 & 7.66 & 0.09 & 0.38 & 3.94 & 69.69 & 66.22 & 65.35 & 17.3 & 14.1\\
  HRED-XL & 8.8 M & 5.08 & 20.45 & 10.90 & 7.22 & 0.11 & 0.50 & 5.51 & 68.25 & 65.93 & 65.09 & 15.9 & 16.9\\
  {\ourmodel} & 7.8 M       & {\bf 7.48} & {\bf 24.38} & {\bf 14.44} & {\bf 10.12} & {\bf 0.17} & {\bf 0.99} & {\bf 25.98} & 86.21 & {\bf 83.18} & 83.76 & 14.3 & 15.0\\
  \hline
\end{tabular}
\caption{{\bf Dialogue} generation performance. All results are averaged over 5 random seeds and reported in percentage (\%). {\bf Size}: Total number of model parameters, {\bf P}: Precision, {\bf R}: Recall, {\bf L-D}: Average dialogue length, {\bf L-U}: Average utterance length. 
  }
\label{table:diagen}
\end{table*}

\begin{table*}[t!]
\fontsize{7.0}{8.4}\selectfont
\centering
\begin{tabular}{p{.148\textwidth}|p{.26\textwidth}| p{.258\textwidth} | p{.245\textwidth} }
  \centering {\bf \fontsize{8}{9.5}\selectfont Goals} & \centering {\bf \fontsize{8}{9.5}\selectfont LM + Goal} & \centering {\bf \fontsize{8}{9.5}\selectfont HRED} &  \multicolumn{1}{p{.24\textwidth}}{\centering {\bf \fontsize{8}{9.5}\selectfont \ourmodel}} \\
  \hline
  Domain: attraction\newline User: [name]\newline  Reqt: [{\bf entrance fee}]\newline Book: False\newline\newline Domain: restaurant\newline User: [area, food]\newline  Reqt: []\newline {\bf Book: True}\newline\newline Domain: {\bf taxi}\newline User: [arriveBy]\newline  Reqt: [car type, {\bf phone}]\newline Book: False\newline
  &  U: what is the address for [attraction\_name] ?\newline S: sure ! [attraction\_name] is located at [attraction\_address] and {\bf is free .}\newline U: i . i am also {\bf looking for a place to eat .} the [value\_area] .\newline S: there are [value\_count] restaurant -s in the [value\_area] . what type of food are you looking for ?\newline ...\newline U: yes , {\bf please book a table} for [value\_count] people at [value\_time] on [value\_day] .\newline ...\newline \underline{\bf U: i help you with anything else ?}\newline S: i {\bf also need a taxi} to get me the restaurant to the restaurant .\newline U: i time would you like to leave ?\newline S: i would like to arrive by the restaurant by the booked time .\newline U: booking completed ! booked car type [taxi\_type] {\bf contact number [taxi\_phone]}\newline S: thank you that s all i needed .\newline U: you are welcome . have a great day !
  & U: what is the address for [attraction\_name] ?\newline S: the address is [hospital\_address] , [value\_place] . is there anything else i can help you with ?\newline U: yes , i would like to book a table for [value\_count] at [value\_time] on [value\_day] .\newline S: i {\bf have booked your table .} your reference number is [restaurant\_reference] .\newline U: thank you . i also need {\bf a train} to [value\_place] that leaves after [value\_time] .\newline S: there are [value\_count] trains that match your request . what day and time would you like to travel ?\newline U: i would like to leave on [value\_day] and arrive by [value\_time] .\newline S: there are [value\_count] trains that meet your criteria . would you like to narrow it down by departure time or arrival time ?\newline U: i would like to arrive by [value\_time] .\newline ...
  & U: what is the address for [attraction\_name] ?\newline S: the address is [attraction\_address] and the postcode is [attraction\_postcode] .\newline U: {\bf what is the entrance fee ?}\newline S: the {\bf entrance fee is free }. is there anything else i can help you with ?\newline U: yes , i am also {\bf looking for a restaurant} that serves [value\_food] food in the [value\_area] .\newline ...\newline S: i have [value\_count] restaurant -s that meet your criteria . would you like me to {\bf book you a table ?}\newline {\bf  U: yes , please .} i need a table for [value\_count] people at [value\_time] on [value\_day] .\newline ...\newline U: i {\bf also need a taxi} to commute between the [value\_count] place -s .\newline S: i have booked you a taxi to pick you up at the [attraction\_name] . it will be a [taxi\_type] and the {\bf contact number is [taxi\_phone]} .\newline U: thank you , that is all i need .\newline S: you are welcome . have a great day !
  \\
  \hline

\end{tabular}
\caption{Dialogue qualitative comparison. Reqt: Requests. U: User, S: Agent. {\bf Goal hit or miss}. \underline{\bf Role confusion}. Extensive qualitative comparisons of dialogues are presented in the appendix. 
  }
\label{table:dialsample}
\end{table*}

\subsection{Evaluation Metrics}
We employ a number of automatic metrics as well as human evaluations to benchmark competing models on quality, diversity, and goal focus:

{\bf Quality.} BLEU \cite{papineni2002bleu}, as BLEU-4 by default, is a word-overlap measure against references and commonly used by dialogue generation works to evaluate quality \shortcite{Sordoni2015ANN, Li2016DeepRL, Li2016ADO, Zhao2017LearningDD, Xu2018BetterCB}. Lower N-gram B1, B2, B3 are also reported.

{\bf Diversity.} D-1, D-2, D-U: The distinctiveness denotes the number of unique unigrams, bigrams, and utterances normalized by each total count \cite{Li2016ADO, Xu2018BetterCB}. These metrics are commonly used to evaluate the dialogue diversity. 

{\bf Goal Focus.} A set of slots such as address are extracted from reference dialogues as multi-label targets.
Generated slots in model's output dialogues are the predictions.
We use the multi-label precision, recall, and F1-score as surrogates to measure the goal focus and achievement.

{\bf Human Evaluation.} The side-by-side human preference  study evaluates dialogues on \emph{goal focus}, \emph{grammar}, \emph{natural flow}, and \emph{non-redundancy}.

\section{Results and Discussion}
\subsection{Dialogue Generation Results}
For dialogue generation \cite{Li2016DeepRL}, a model is given one or more goals and one user utterance as the seed inputs to generate entire dialogues in an auto-regressive manner.

Table \ref{table:diagen} summarizes the evaluation results.
For quality measures, {\ourmodel} significantly outperforms other baselines, implying that it's able to carry out a higher-quality dialogue.
Besides, goal-embedded LMs perform better than HREDs, showing the benefits of our goal embedding module.
No significant performance difference is observed with respect to model size variants.

\begin{table*}[t!]
\centering
\smaller[1]
\begin{tabular}{l|c c c c | c c c | c c c | c}
  \hline
  {\bf Model} & {\bf BLEU} & {\bf B1}  & {\bf B2}  & {\bf B3}  & {\bf D-1}  & {\bf D-2}  & {\bf D-U} & {\bf P} & {\bf R} & {\bf F1} & {\bf L-R}\\
  \hline
  \hline
  LM+G    & 14.88 & 35.86 & 24.59 & 18.81 & 0.27 & 1.44 & 40.84 & 79.71 & 68.57 & 71.73 & 14.3 \\
  LM+G-XL & 14.51 & 35.28 & 24.07 & 18.36 & {\bf 0.28} & {\bf 1.47} & {\bf 42.56} & {\bf 79.79} & 67.31 & 71.00 & 14.3 \\
  HRED    & 14.34 & 36.27 & 24.31 & 18.33 & 0.21 & 0.94 & 20.21 & 75.46 & 67.08 & 68.78 & 17.1 \\
  HRED-XL & 14.33 & 36.36 & 24.37 & 18.35 & 0.23 & 1.12 & 26.63 & 72.69 & 68.24 & 68.20 & 17.3 \\
  {\ourmodel}   & {\bf 15.85} & {\bf 37.99} & {\bf 26.14} & {\bf 20.01} & 0.25 & 1.27 & {\bf 39.59} & {\bf 78.34} & {\bf 71.55} & {\bf 72.69} & 16.7\\
  \hline
\end{tabular}
\caption{{\bf Agent's response} generation performance. All results are  averaged over 5 random seeds and reported in percentage (\%). {\bf P}: Precision, {\bf R}: Recall, {\bf L-R}: Average response length. 
  }
\label{table:resgensys}
\end{table*}

\begin{table*}[t!]
\centering
\smaller[1]
\begin{tabular}{l|c c c c | c c c | c c c | c}
  \hline
  {\bf Model} & {\bf BLEU} & {\bf B1}  & {\bf B2}  & {\bf B3}  & {\bf D-1}  & {\bf D-2}  & {\bf D-U} & {\bf P} & {\bf R} & {\bf F1} & {\bf L-R}\\
  \hline
  \hline
  LM+G    & 11.73 & 31.79 & 21.26 & 15.56 & 0.35 & 1.82 & 33.57 & 89.44 & 75.78 & 80.23 & 10.6 \\
  LM+G-XL & 11.60 & 31.49 & 21.00 & 15.38 & {\bf 0.36} & {\bf 1.87} & 34.29 & 89.57 & 75.55 & 80.03 & 10.7 \\
  HRED    & 10.88 & 31.69 & 20.46 & 14.65 & 0.24 & 0.98 & 16.00 & 80.00 & 79.11 & 77.58 & 13.1 \\
  HRED-XL & 10.81 & 31.84 & 20.48 & 14.60 & 0.26 & 1.15 & 19.87 & 80.11 & 78.82 & 77.42 & 13.2 \\
  {\ourmodel}  & {\bf 13.25} & {\bf 35.20} & {\bf 23.89} & {\bf 17.56} & 0.30 & 1.49 & {\bf 35.57} & {\bf 91.12} & {\bf 79.66} & {\bf 83.51} & 12.8 \\
  \hline
\end{tabular}
\caption{{\bf User's response} generation performance. All results are  averaged over 5 random seeds and reported in percentage (\%). {\bf P}: Precision, {\bf R}: Recall, {\bf L-R}: Average response length. 
  }
\label{table:resgenusr}
\end{table*}

\begin{table*}[t!]
\fontsize{8}{9.5}\selectfont
\centering
\begin{tabular}{p{.16\textwidth} |p{.44\textwidth} | p{.32\textwidth} }
  \centering {\bf Goals} & \centering {\bf Contexts (Reference)} &  \multicolumn{1}{p{.32\textwidth}}{\centering {\bf Response}} \\
  \hline
  Domain: train\newline User: [destination, departure, arriveBy, day]\newline Reqt:[duration,price]\newline Book: False\newline\newline Domain: restaurant\newline User: [area, food, pricerange]\newline  Reqt: []\newline Book: True
  &  U: hey i am looking for a train from [value\_place] to [value\_place] . mind helping out ?\newline S: many trains are leaving on [value\_day] . mind choosing the time ?\newline U: i want to leave on [value\_day] and arrive by [value\_time] .\newline S: i have train [train\_id] that leaves [value\_place] at [value\_time] and arrives in [value\_place] at [value\_time] . would you like to make reservations ?\newline U: could you give me the travel time and price of that train please ?\newline S: the cost is [value\_price] pounds . the duration of the trip is [value\_count] minutes , leaving [value\_place] at [value\_time] and arriving at [value\_time] in [value\_place] .
  & {\bf LM+G}: thanks . i am also looking for a restaurant in the [value\_area] .\newline{\bf HRED}: i am also looking for a place to stay . i would like a [value\_count] star hotel .\newline {\bf {\ourmodel}}: thank you . i am also looking for a restaurant in the [value\_area] of town .
  \\
  \hline
  Domain: train\newline User: [destination, departure, arriveBy, day]\newline  Reqt: []\newline Book: False
  & U: i need a train that leaves on [value\_day] with an arrival of [value\_time] .\newline S: where will you be travelling to and from ?
  & {\bf LM+G}: i will be departing from [value\_place] . \newline{\bf HRED}: i will be leaving from [value\_place] . \newline {\bf {\ourmodel}}: i am departing from [value\_place] and need to arrive by [value\_time] .
  \\
  \hline
\end{tabular}
\caption{Qualitative comparison of generated responses. Reqt: Requests. U: User, S: Agent.
  }
\label{table:respsample}
\end{table*}

For diversity evaluations, {\ourmodel} is on par with goal-embedded LMs and both outperform HRED significantly.
Of $1000$ generated dialogues, HRED delivers highly repetitive outputs with only $4$ to $6\%$ distinct utterances, whereas $25\%$ of utterances are unique from {\ourmodel}.

For recovering slots in reference dialogues, precision denotes a degree of goal deviation, recall entails the achievement of goals, and F1 measures the overall focus.
Goal-embedded LM is the best on precision and F1 with {\ourmodel} having comparable performance.
However, even thought LM can better mention the slots in dialogue generation, utterances are often associated with a wrong role.
That is, role confusions are commonly seen such as the user makes reservations for the agent as in Table \ref{table:dialsample}.
The reason could be that LM handles the task similar to paragraph generation without an explicit design for the conversational hierarchy.

Overall, {\ourmodel} is able to generate high-quality dialogues with sufficient diversity and still adhere to goals compared to baselines.

{\bf Qualitative Comparison.} 
Table \ref{table:dialsample} compares generated dialogues from different models given one to three goals to focus on.
It's clear that models with the goal embedding module are able to adhere to given goals such as ``book" or ``no book", requesting ``price" or ``entrance fee" while HRED fails to do so.
They can also correctly covering all required domain contexts without any diversion such as switching from attraction inquiry to restaurant booking, then to taxi-calling.
For HRED, without goals, generated dialogues often detour to a non-relevant domain context such as shifting to train booking while only hotel inquiry is required.

For goal-embedded LM, a serious issue revealed is role confusions as LM often wrongly shifts between the user and agent as shown in Table \ref{table:dialsample}.
The issue results from one wrong \texttt{EndofUtterance} prediction but affects rest of dialogue and degrades the overall quality.
More generated dialogues are reported in the appendix.

\subsection{Response Generation Results}
\vspace{-0.1mm}
For response generation \cite{Sordoni2015ANN, Serban2016AHL, Park2018AHL}, a set of goals as well as the previous context, i.e. all previous reference utterances, are given to a model to generate the next utterance as a response.

Table \ref{table:resgensys} and \ref{table:resgenusr} summarize the results.
{\ourmodel} outperforms others on quality and goal focus measures and rivals LM-goal on diversity on both agent and user responses.
For goal focus, LM-goal performs good on precision but short on recall. This could because it generates much shorter user and agent responses on average. 

Interestingly, as previous contexts are given, LM-goal performs only slightly better than HRED.
This implies hierarchical structures capturing longer dependencies can make up the disadvantages of having no goal information for response generation.
However, as illustrated in Table \ref{table:respsample}, HRED could still fail to predict the switch of domain contexts, e.g. from \texttt{train} to \texttt{restaurant}, which explains performance gaps.
Another intriguing observation is that when incorporating the goal embedding module, response diversity and goal focus can be boosted significantly.

Comparing the performance between agent and user response generation, we observe that models can achieve higher quality and diversity but lower goal focus when modeling agent's responses.
These might result from the relatively consistent utterance patterns but diverse slot types used by an agent. 
More generated responses across different models are presented in the appendix.

\begin{table}[t!]
\fontsize{10.5}{12.5}\selectfont
\centering
\begin{tabular}{l|c c c}
  \hline
   & {\bf Wins} & {\bf Losses} & {\bf Ties} \\
  \hline
  Goal Focus & {\bf 82.33\%} & 6.00\% & 11.67\% \\[0.5mm]
  Grammar & 6.00\% & 5.00\% & 89.00\% \\[0.5mm]
  Natural Flow & {\bf 26.00\%} & 15.00\% & 59.00\% \\[0.5mm]
  Non-redundancy  & {\bf 35.34\%} & 6.33\% & 58.33\% \\[0.5mm]
  \hline
\end{tabular}
\caption{Human evaluations, {\ourmodel} vs HRED. $100$ pairs of generated dialogues along with goals are given to three domain experts for side-by-side comparisons.
  }
\label{table:humaneval}
\end{table}

\subsection{Human Evaluation Results}
For human evaluation, we conduct side-by-side comparisons between {\ourmodel} and HRED, the widely used baseline in literature, on dialogue generation task.
We consider the following four criteria: \emph{goal focus}, \emph{grammar}, \emph{natural flow}, and \emph{non-redundancy}.
\emph{Goal focus} evaluates whether the dialogue is closely related to the preset goals; \emph{grammar} evaluates whether the utterances are well-formed and understandable; \emph{natural flow} evaluates whether the flow
of dialogue is logical and fluent; and \emph{non-redundancy} evaluates whether the dialogue is absent of unnecessary repetition of mentioned information.
$100$ pairs of generated dialogues from {\ourmodel} and HRED along with their goals are randomly placed against each other.
For each goal and pair of dialogues, three domain experts were instructed to set their preferences with respect to each of the four criteria, marked as \emph{win / lose / tie} between the dialogues.

Table \ref{table:humaneval} presents the results.
{\ourmodel} shows substantial advantages on goal focus, with $82.33\%$ wins over HRED, confirming the benefits of our goal embedding module.
{\ourmodel} also outperforms HRED significantly on natural flow and non-redundancy.
These might result from {\ourmodel}'s ability to generating much more diverse utterances while concentrating on current goals. 
An especially interesting observation is that in cases where multiple goals are given, {\ourmodel} not only stays focused on each individual goal but also generates intuitive transitions between goals, so that the flow of a dialogue is natural and coherent. 
An example is shown in Table \ref{table:dialsample}, where the {\ourmodel}-generated dialogue switches towards the \texttt{taxi} goal while maintaining reference to the previously mentioned \texttt{attraction} and \texttt{restaurant} goals: \textit{``\ldots i also need a taxi to commute between the 2 places \ldots''}. We also observe that both {\ourmodel} and HRED performed well on grammaticality. The generated samples across all RNN-based models are almost free from grammar error as well.

\begin{table*}[t!]
\centering
\begin{tabular}{l|c c | c c c | c c c}
  \hline
  {\bf Model} & {\bf BLEU} & {\bf B1} & {\bf D-1}  & {\bf D-2}  & {\bf D-U} & {\bf P} & {\bf R} & {\bf F1}\\
  \hline
  \hline
  {\ourmodel} & 7.48 & 24.38 & 0.17 & 0.99 & 25.98 & 86.21 & 83.18 & 83.76\\
  \quad w/o goal & 5.19 & 20.04 & 0.13 & 0.68 & 13.83 & 69.18 & 68.21 & 66.86\\
  \quad w/o dual & 7.34 & 24.99 & 0.15 & 0.79 & 19.22 & 85.24 & 82.62 & 82.96\\
  \quad w/o context attention & 7.34 & 24.34 & 0.17 & 0.99 & 24.98 & 86.70 & 83.40 & 84.10\\
  \hline
\end{tabular}
\caption{Ablation studies on {\bf dialogue} generation over goal embedding module, dual architecture, and dialogue-level attention. Results are averaged over 5 random seeds and reported in percentage (\%). {\bf P}: Precision, {\bf R}: Recall.
  }
\label{table:dialabl}
\end{table*}

\begin{table*}[t!]
\centering
\begin{tabular}{l|c c | c c c | c c c}
  \hline
  {\bf Model} & {\bf BLEU} & {\bf B1} & {\bf D-1}  & {\bf D-2}  & {\bf D-U} & {\bf P} & {\bf R} & {\bf F1}\\
  \hline
  \hline
  {\ourmodel} & 14.84 & 36.84 & 0.18 & 1.10 & 34.23 & 89.87 & 82.00 & 84.80\\
  \quad w/o goal & 13.29 & 35.23 & 0.16 & 0.97 & 28.04 & 84.33 & 80.15 & 81.10\\
  \quad w/o dual & 14.60 & 36.66 & 0.17 & 0.96 & 27.53 & 89.43 & 80.81 & 83.91\\
  \quad w/o context attention & 14.73 & 36.88 & 0.18 & 1.14 & 34.55 & 90.28 & 81.54 & 84.80\\
  \hline
\end{tabular}
\caption{Ablation studies on {\bf response} generation over goal embedding module, dual architecture, and dialogue-level attention. Results are averaged over 5 random seeds and reported in percentage (\%). {\bf P}: Precision, {\bf R}: Recall.
  }
\label{table:respabl}
\end{table*}

\subsection{Ablation Studies}

The ablation studies are reported in Table \ref{table:dialabl} for dialogue generation and in Table \ref{table:respabl} for response generation to investigate the contribution of each module.
Here we evaluate user and agent response generation together.

\vspace{0.5mm}
\noindent
{\bf Goal Embedding Module.} First, we examine the impact of goal embedding module.
When unplugging the goal embedding module, we observe significant and consistent drops on quality, diversity, and goal focus measures for both dialogue and response generation tasks.
For dialogue generation task, the drops are substantially large which resonates with our intuition as the model only has the first user utterance as input context to follow.

With no guideline about what to achieve and what conversation flow to go around with, dialogues generated from HRED often have the similar flow and low diversity.
These results demonstrate that our goal embedding module is critical in generating higher-quality and goal-centric dialogues with much more diversity.

\vspace{0.5mm}
\noindent
{\bf Dual Hierarchical Architecture.} We also evaluate the impact of dual hierarchical architecture.
Comparisons on both dialogue and response generation tasks show a consistent trend.
We observe that applying dual architecture for interlocutor-level modeling leads to a solid increase in utterance diversity as well as moderate improvements on quality and goal focus.

The results echo our motivation as two interlocutors in a goal-oriented dialogue scenario exhibit distinct conversational patterns and this interlocutor-level disparity should be modeled by separate hierarchical encoder-decoders.

For the dialogue-level attention module, there is no significant effect on diversity and goal focus on both tasks but it marginally improves the overall utterance quality as BLEU scores go up by a bit.

\begin{table}[t!]
\centering
\begin{tabular}{c|c c}
  \hline
    & Joint Goal & Turn Request    \\
  \hline
  GLAD & 88.55\% & 97.11\% \\
  GLAD + LM+G & 88.07\% & 96.02\% \\
  GLAD + HRED & 89.03\% & 97.11\% \\
  GLAD + {\ourmodel} & {\bf 89.04\%} & {\bf 97.59\%}$^{\ast}$ \\
  \hline
\end{tabular}
\caption{Test accuracy of GLAD \cite{Zhong2018GlobalLocallySD} on the WoZ restaurant reservation dataset with different data augmentation models. ($^{\ast}$significant against others.) 
  }
\label{table:augment}
\end{table}

\section{Data Augmentation via Dialogue Generation}
As an exemplified extrinsic evaluation, we leverage the goal-oriented dialogue generation as data augmentation for task-oriented dialogue systems.
Dialogue state tracking (DST) is used as our evaluation task which is a critical component in task-oriented dialogue systems \cite{Young2013POMDPBasedSS} and has been studied extensively \cite{henderson2014word, Mrksic2017NeuralBT, Zhong2018GlobalLocallySD}.

In DST, given the current utterance and dialogue history, a dialogue state tracker determines the state of the dialogue which comprises a set of {\it requests} and {\it joint goals}.
For each user turn, the user informs the system a set of  turn goals to fulfill, {\it e.g. inform(area=south)}, or turn requests asking for more information, {\it e.g. request(phone)}.
The joint goal is the collection of all turn goals up to the current turn.

We use the state-of-the-art Global-Locally Self-Attentive Dialogue State Tracker (GLAD) \cite{Zhong2018GlobalLocallySD} as our benchmark model and the WoZ restaurant reservation dataset \cite{Wen2017ANE, Zhong2018GlobalLocallySD} as our benchmark dataset, which is commonly used for the DST task.

The dataset consists of $600$ train, $200$ validation and $400$ test dialogues.
We use the first utterances from $300$ train dialogues and sample restaurant-domain goals to generate dialogues, whose states are annotated by a rule-based method.

Table \ref{table:augment} summarizes the augmentation results.
Augmentation with {\ourmodel} achieved an improvement over the vanilla dataset and outperform HRED on turn requests while being comparable on joint goal.
For goal-embedded LM, as it struggles with role confusion, the augmentation actually hurts the overall performance.

\section{Conclusion}
We introduced the goal-embedded dual hierarchical attentional encoder-decoder ({\ourmodel}) for goal-oriented dialogue generation.
{\ourmodel} is able to generate higher-quality and goal-focused dialogues as well as responses with decent diversity and non-redundancy.
Empirical results show that the goal embedding module plays a vital role in the performance improvement and the dual architecture can significantly enhance diversity. 

We demonstrated one application of the goal-oriented dialogue generation through a data augmentation experiment, though the proposed model is applicable to other conversational AI tasks which remains to be investigated in the future.

As shown in experiments, a language model coupled with goal embedding suffers from role-switching or confusion. It's also interesting to further dive deep with visualizations \cite{kessler2017scattertext} and quantify the impact on quality, diversity, and goal focus metrics.

\section*{Acknowledgments}
The authors would like to acknowledge the entire AWS Lex Science team for thoughtful discussions, honest feedback, and full support.
We are also very grateful to the reviewers for insightful comments and helpful suggestions.

\bibliography{conll-2019}
\bibliographystyle{acl_natbib}

\clearpage
\input{appendix.tex}

\end{document}

%% file: appendix.tex
%
%









\appendix

\section{Appendix}

\subsection{More Generated Dialogues for Qualitative Comparison}
\label{sec:dialsample}
In Table \ref{table:dialsamp} and Table \ref{table:dialsamp2}, we list and compare more generated dialogues of different models given a variable number of goals to focus on.

The domains, what slots to request such as phone or postcode, and when should the user make a reservation or not are specified in goals.
We use bold style to emphasize whether the assigned goals are achieved or missed such as requests of a correct slot or a wrong domain.
The underline and bold style denotes the role confusions.

\begin{table*}[b]
\fontsize{7.0}{8.4}\selectfont
\centering
\begin{tabular}{p{.15\textwidth} |p{.25\textwidth} | p{.255\textwidth} | p{.25\textwidth} }
  \centering {\bf \fontsize{8}{9.5}\selectfont Goals} & \centering {\bf \fontsize{8}{9.5}\selectfont LM + Goal} & \centering {\bf \fontsize{8}{9.5}\selectfont HRED} &  \multicolumn{1}{p{.25\textwidth}}{\centering {\bf \fontsize{8}{9.5}\selectfont {\ourmodel}}} \\
  \hline
  Domain: {\bf hotel}\newline User: [name]\newline  Reqt: []\newline {\bf Book: True}\newline
  &  U: i am looking for a hotel called the [hotel\_name] .\newline S: [hotel\_name] is a [value\_pricerange] -ly priced guesthouse in the [value\_area] . would you like me to book a room for you ?\newline U: yes , please . i need [value\_count] for [value\_count] people and [value\_count] nights starting from [value\_day] .\newline S: {\bf i have booked you at the [hotel\_name] .} your reference number is [hotel\_reference] . is there anything else i can help you with ?\newline U: no , that s all . thank you .\newline S: you are welcome . have a great day !\newline
  & U: i am looking for a hotel called the [hotel\_name] .\newline S: [hotel\_name] is a guesthouse in the [value\_area] . it is [value\_pricerange] -ly priced and has [value\_count] stars . would you like to book a room ?\newline U: yes , please . i need a room for [value\_count] people for [value\_count] nights starting on [value\_day] .\newline S: i am sorry , but i was unable to book that for you . would you like to try a different day or a shorter stay ?\newline U: how about [value\_count] nights ?\newline S: {\bf i was able to book you at the [hotel\_name] .} your reference number is [hotel\_reference] . is there anything else i can help you with ?\newline U: {\bf i am also looking for a train} that leaves after [value\_time] on [value\_day] .\newline ... (Cut Off) $\rightarrow$ {\bf Off the Rails}\newline
  & U: i am looking for a hotel called the [hotel\_name] .\newline S: [hotel\_name] is a guesthouse in the [value\_area] area . would you like to book a room ?\newline U: yes , i would like to book it for [value\_count] people and [value\_count] nights starting from [value\_day] .\newline S: {\bf booking was successful .} reference number is [hotel\_reference] .\newline U: that s all i need . thank you .\newline S: you are welcome . have a great day !\newline
  \\
  \hline
  Domain: attraction\newline User: [name]\newline  Reqt: [{\bf phone}, type]\newline Book: False\newline\newline Domain: train\newline User: [arriveBy, day, departure, destination]\newline  Reqt: [{\bf price}]\newline {\bf Book: False}\newline
  & U: i am looking for information about [attraction\_name]\newline S: [attraction\_name] is a nightclub in the [value\_area] of town . their {\bf phone number is [attraction\_phone]} .\newline \underline{\bf U: i help you with anything else ?}\newline S: yes , i am also looking for a train that [value\_place] to [value\_place] .\newline U: what day and time would you like to leave ?\newline S: i would like to leave on [value\_day] and arrive by [value\_time] .\newline U: i have [value\_count] trains that that day . [value\_time] . would you like me to book it for you ?\newline S: yes , please . {\bf i need the price} and train id .\newline U: the train is [value\_price] pounds . the price is [value\_price] pounds .\newline S: you like me to {\bf book that for you ?}\newline {\bf U: no} , that s all i need . thank you .\newline S: thank you for using our service . have a great day .\newline
  &  U: i am looking for information about [attraction\_name]\newline S: [attraction\_name] is located in the [value\_area] of town at [attraction\_address] .\newline U: what is the entrance fee ?\newline S: i am sorry , i do not have that information .\newline U: i am also looking for a train that leaves after [value\_time] on [value\_day] .\newline S: i have [value\_count] trains that meet your criteria . would you like to narrow it down by departure site , destination , or arrival time ?\newline U: i would like to leave on [value\_day] and arrive by [value\_time] .\newline S: i have train [train\_id] that leaves at [value\_time] and arrives at [value\_time] . would you like me to book that for you ?\newline U: {\bf yes , please book it} for [value\_count] people .\newline S: booking was successful , the total fee is [value\_price] gbp payable at the station . reference number is [train\_reference] .\newline U: thank you . that is all i need .\newline S: you are welcome . have a great day !  \newline
  & U: i am looking for information about [attraction\_name]\newline S: [attraction\_name] is a museum in the [value\_area] area . would you like the phone number ?\newline U: yes , i would like the postcode and phone number .\newline S: their {\bf phone number is  [attraction\_phone]} and the postcode is [attraction\_postcode] .\newline U: i also need a train to [value\_place] .\newline S: where will you be departing from ?\newline U: i will be leaving from [value\_place] on [value\_day] .\newline S: i have [value\_count] trains that meet your criteria . what time would you like to arrive by ?\newline U: i need to arrive by [value\_time] .\newline S: i have train [train\_id] that leaves at [value\_time] and arrives at [value\_time] . would you like me to {\bf book that for you ?}\newline {\bf U: no }, that s all i needed . thank you .\newline S: you are welcome . have a great day !\newline
  \\
  \hline
\end{tabular}
\caption{Dialogue qualitative comparison. Reqt: Requests. U: User, S: System. {\bf Goal hit or miss}. \underline{\bf Role confusion}.
  }
\label{table:dialsamp}
\end{table*}



\begin{table*}[p!]
\fontsize{7.0}{8.4}\selectfont
\centering
\begin{tabular}{p{.15\textwidth} |p{.25\textwidth} | p{.255\textwidth} | p{.25\textwidth} }
  \centering {\bf \fontsize{8}{9.5}\selectfont Goals} & \centering {\bf \fontsize{8}{9.5}\selectfont LM + Goal} & \centering {\bf \fontsize{8}{9.5}\selectfont HRED} &  \multicolumn{1}{p{.25\textwidth}}{\centering {\bf \fontsize{8}{9.5}\selectfont {\ourmodel}}} \\
  \hline
  Domain: attraction\newline User: [name]\newline  Reqt: [address, entrance fee]\newline Book: False\newline\newline Domain: {\bf hotel}\newline User: [name]\newline  Reqt: [postcode, stars]\newline {\bf Book: False}\newline\newline Domain: {\bf taxi}\newline User: [leaveAt]\newline  Reqt: [car type, phone]\newline Book: False\newline 
  & U: i am looking for [attraction\_name] in [value\_place]\newline S: it is located in the [value\_area] area . it is located on [attraction\_address] .\newline \underline{\bf U: is free .}\newline \underline{\bf S: can i get the phone number ?}\newline U: the , the phone number is [attraction\_phone] .\newline S: i you . {\bf i am also looking for a hotel called [hotel\_name] .}\newline U: [hotel\_name] is a [value\_pricerange] hotel in the [value\_area] area . it is a [value\_count] star rating . would you like me to {\bf book a room for you ?\newline S: no ,} i just need the information .\newline U: you are welcome . is there anything else i can help you with today ?\newline S: {\bf i also need a taxi} to commute between the [value\_count] place -s .\newline U: what time would you like to leave ?\newline S: i would like to leave the hotel by [value\_time] .\newline U: i have booked a taxi for you . the contact number is [taxi\_phone] . is there anything else i can help you with ?\newline S: no , that s all i need . thank you for your help !\newline U: you are welcome . have a great day !\newline
  & U: i am looking for [attraction\_name] in [value\_place]\newline S: i have found [attraction\_name] , it s located in the [value\_area] . it is free to enter . would you like more information ?\newline U: yes , can i get the phone number and postcode ?\newline S: sure , the phone number is [attraction\_phone] and the postcode is [attraction\_postcode] .\newline {\bf U: i am also looking for a train} that leaves on [value\_day] and goes to [value\_place] .\newline S: there are [value\_count] trains that match your request . what time would you like to leave ?\newline U: i would like to leave after [value\_time] .\newline S: i have train [train\_id] that leaves at [value\_time] and arrives at [value\_time] . would you like me to book that for you ?\newline U: yes , please book it for [value\_count] people .\newline S: booking was successful , the total fee is [value\_price] gbp payable at the station . reference number is [train\_reference] .\newline U: thank you so much for your help .\newline S: you are welcome . have a great day !\newline U: thank you , you too .\newline S: thank you for using our service today ! \newline $\rightarrow$ {\bf Off the Rails}\newline
  & U: i am looking for [attraction\_name] in [value\_place]\newline S: it is located at [attraction\_address] . is there anything else you would like to know ?\newline U: what is the postcode ?\newline S: the postcode is [hotel\_postcode] . is there anything else i can help you with ?\newline {\bf U: i am also looking for a hotel called [hotel\_name] .}\newline S: [hotel\_name] is a [value\_pricerange] -ly priced guesthouse in the [value\_area] area . would you like me to {\bf book you a room ?\newline U: no , }i just need the address please .\newline S: the postcode is [hotel\_postcode] . is there anything else i can help you with ?\newline {\bf U: i also need a taxi} to get between the [value\_count] place -s .\newline S: i can help you with that . what time would you like to leave or arrive by ?\newline U: i want to leave the hotel by [value\_time] .\newline S: i have booked you a [taxi\_type] , the contact number is [taxi\_phone] .\newline U: thank you so much for your help .\newline S: you are welcome . is there anything else i can help you with today ?\newline U: no , that s all i need . thank you for your help !\newline S: you are welcome . have a great day . \newline
  \\
  \hline
  Domain: hotel\newline User: [name]\newline  Reqt: [address]\newline {\bf Book: False}\newline\newline Domain: {\bf attraction}\newline User: [area]\newline  Reqt: [address, phone, type]\newline Book: False\newline
  & U: hello ! i am looking for information about [hotel\_name] . it s a hotel in [value\_place] .\newline S: the is a guesthouse in the [value\_area] part of town . it is [value\_pricerange] -ly priced and has [value\_count] stars . would you like me to book a room ?\newline {\bf U: no , i do not need a reservation .} i you .\newline S: you are welcome . is there anything else i can help you with today ?\newline U: yes , i am {\bf looking for a place called [attraction\_name] .}\newline S: [attraction\_name] is a museum in the [value\_area] area . it is located at [attraction\_address] . their phone number is [attraction\_phone] .\newline \underline{\bf U: i help you with anything else ?}\newline S: no , that s all i need . thank you .\newline U: you are welcome . have a great day !\newline
  & U: hello ! i am looking for information about [hotel\_name] . it s a hotel in [value\_place] .\newline S: [hotel\_name] is a guesthouse in the [value\_area] . it is [value\_pricerange] -ly priced and has [value\_count] stars . would you like to book a room ?\newline U: yes , please . i need a room for [value\_count] people for [value\_count] nights starting on [value\_day] .\newline S: i am sorry , but i was unable to book that for you . would you like to try a different day or shorter stay ?\newline U: how about [value\_count] nights ?\newline {\bf S: i was able to book you} at the [hotel\_name] . your reference number is [hotel\_reference] . is there anything else i can help you with ?\newline {\bf U: i am also looking for a train} that leaves after [value\_time] on [value\_day] .\newline ... (Cut Off) $\rightarrow$ {\bf Off the Rails}\newline
  & U: hello ! i am looking for information about [hotel\_name] . it s a hotel in [value\_place] .\newline S: [hotel\_name] is a guesthouse in the [value\_area] area . it is a [value\_count] star hotel in the [value\_area] area . would you like me to book a room for you ?\newline {\bf U: no , i just need the address} and phone number .\newline S: sure , the phone number is [attraction\_phone] .\newline {\bf U: i am also looking for a place to go in town} .\newline S: there are [value\_count] attractions in the [value\_area] area . do you have a particular type of attraction in mind ?\newline U: i am not sure . i just need the address and postcode for [value\_count] of them .\newline S: [attraction\_name] is located at [attraction\_address] and the postcode is [attraction\_postcode] .\newline U: no , that s all i need . thank you for your help !\newline S: you are welcome . have a great day !\newline
  \\
  \hline
  Domain: train\newline User: [day, departure, destination, leaveAt]\newline  Reqt: [{\bf duration, price}]\newline {\bf Book: False}\newline
  & U: i am looking for a train from [value\_place] to [value\_place] .\newline S: there are [value\_count] trains that meet your criteria . what day and time would you like to travel ?\newline U: i would like to leave on [value\_day] after [value\_time] .\newline S: i have [value\_count] trains that leave [value\_place] at [value\_time] . would you like me to {\bf book it for you ?\newline U: yes ,} please . i need {\bf the price and travel time please .}\newline S: the price is [value\_price] pounds . the price is [value\_price] pounds .\newline \underline{\bf U: you like me to book that for you ?}\newline S: no , that s all i need . thank you .\newline U: you are welcome . have a great day !\newline
  & U: i am looking for a train from [value\_place] to [value\_place] .\newline S: there are [value\_count] trains that meet your criteria . what day and time would you like to travel ?\newline U: i would like to leave on [value\_day] and arrive by [value\_time] .\newline S: i have train [train\_id] that leaves at [value\_time] and arrives in [value\_place] at [value\_time] . would you like me to {\bf book that for you ?\newline U: yes ,} please . i need [value\_count] tickets .\newline S: i have booked [value\_count] tickets for you . your reference number is [train\_reference] .\newline U: i am also looking for a place to stay . {\bf i would like a [value\_count] star hotel .}\newline ... (Cut Off) $\rightarrow$ {\bf Off the Rails}\newline
  & U: i am looking for a train from [value\_place] to [value\_place] .\newline S: what day would you like to travel ?\newline U: i would like to leave on [value\_day] .\newline S: there are [value\_count] trains that meet your criteria . do you have a time you would like to leave or arrive by ?\newline U: i would like to leave after [value\_time] .\newline S: [train\_id] leaves at [value\_time] and arrives at [value\_time] . would you like me to {\bf book it for you ?\newline U: no ,} i just need {\bf the price} and the train id .\newline S: the price is [value\_price] pounds .\newline U: thank you for your help . that is all i need today .\newline S: you are welcome . have a great day ! \newline
  \\
  \hline
\end{tabular}
\caption{Dialogue qualitative comparison (continue). Reqt: Requests. U: User, S: System. {\bf Goal hit or miss}. \underline{\bf Role confusion}.
  }
\label{table:dialsamp2}
\end{table*}

\subsection{More Generated Responses for Qualitative Comparison}
\label{sec:respsample}
Responses are generated by a model given the goals and all the previous utterances.
In Table \ref{table:respsample}, we present more generated responses for qualitative comparison.
We observe that HRED is unable to correctly switch beteen goal contexts illustrated by the second example.

\begin{table*}[b]
\fontsize{8}{9.5}\selectfont
\centering
\begin{tabular}{p{.14\textwidth} |p{.48\textwidth} | p{.3\textwidth} }
  \centering {\bf Goals} & \centering {\bf Contexts (Reference)} &  \multicolumn{1}{p{.3\textwidth}}{\centering {\bf Response}} \\
  \hline
  Domain: attraction\newline User: [type]\newline  Reqt: [entrance fee, address, phone]\newline Book: False\newline\newline Domain: restaurant\newline User: [pricerange, food]\newline  Reqt: [phone, postcode]\newline Book: False
  &  U: can you find me a theatre ?\newline S: would you like the [attraction\_name] ?\newline U: yes and i need to know the entrance fee , address , and phone number\newline S: the fee is not listed but the address is [attraction\_address] and the phone number is [attraction\_phone] . can i help you with anything else ?\newline U: i am also looking for a [value\_food] restaurant in the [value\_pricerange] range .\newline S: i have [value\_count] excellent options . [restaurant\_name] is in the [value\_area] of town , would you like me to reserve you a table ?\newline U: i am not sure how many people will be coming , so i will do that later. but can you provide me with their postcode and phone number ?
  & {\bf LM+G}: the phone number is [restaurant\_phone] and the postcode is [restaurant\_postcode] .\newline{\bf HRED}: the phone number is [restaurant\_phone] and the postcode is [restaurant\_postcode] .\newline {\bf {\ourmodel}}: sure , the phone number is [restaurant\_phone] and the postcode is [restaurant\_postcode] . is there anything else i can help you with ?
  \\
  \hline
  Domain: train\newline User: [day, departure, destination, leaveAt]\newline  Reqt: []\newline Book: True\newline\newline Domain: attraction\newline User: [area, type]\newline  Reqt: [phone, postcode]\newline Book: False\newline
  & U: i am looking for a train that leaves on [value\_day] going to [value\_place] .\newline S: there are [value\_count] trains that meet what you are asking . where would you like to depart from ?\newline U: [value\_place] , it should also leave after [value\_time] .\newline S: [train\_id] departs [value\_place] at [value\_time] and arrives in [value\_place] at [value\_time] . will that work ?\newline U: yes it will . can you book that for just me , and provide the reference number ?\newline S: sure thing ! how many tickets do you need ?\newline U: i need [value\_count] ticket and can i please have the reference number ?\newline S: booking was successful , the total fee is [value\_price] gbp payable at the station . reference number is [train\_reference] . can i provide any further help today ?
  & {\bf LM+G}: i am also looking for a place to go in the [value\_area] . \newline{\bf HRED}: i am also looking for a place to stay . i would like a hotel in the [value\_area] . \newline {\bf {\ourmodel}}: i am also looking for a place to go in town .
  \\
  \hline
  Domain: hotel\newline User: [area, internet, parking, stars]\newline  Reqt: [postcode, pricerange, type]\newline Book: False\newline\newline Domain: restaurant\newline User: [area, food, pricerange]\newline  Reqt: []\newline Book: True
  & U: hi i am looking for a restaurant in the [value\_area] of town serving [value\_food] food .\newline S: that s great ! there are several [value\_food] restaurant -s near the city [value\_area] . are you looking for a budget @UNK option or are you comfortable with a more [value\_pricerange] place ?\newline U: a [value\_pricerange] option please .\newline S: we have [restaurant\_name] , the ghandi and [restaurant\_name] , do you want me to book you for [value\_count] of those ?\newline U: yes please for [value\_count] people at [value\_time] on [value\_day].\newline S: you are all set . they will hold the table for [value\_count] minutes and your reference number is [hotel\_reference]\newline U: thank you , appreciate your service .\newline S: no problem . is there anything else i can help you with ?
  & {\bf LM+G}: i am also looking for a place to stay . i would like a [value\_count] star hotel in the [value\_area] . \newline{\bf HRED}: no , that s all . thank you . \newline {\bf {\ourmodel}}: i am also looking for a place to stay .
  \\
  \hline
  Domain: taxi\newline User: [destination, arriveBy, departure]\newline  Reqt: [car type, phone]\newline Book: False
  & U: i need to book a tax departing from gandhi .\newline S: what is your destination and when would you like to depart ?
  & {\bf LM+G}: i need to arrive by [value\_time] . \newline{\bf HRED}: i am going to [value\_place] from [value\_place] . \newline {\bf {\ourmodel}}: i want to go to [value\_place] and get to [restaurant\_name] by [value\_time] .
  \\
  \hline
  Domain: attraction\newline User: [area]\newline  Reqt: [address]\newline Book: False \newline \newline Domain:hotel\newline User: [pricerange, internet, stars]\newline  Reqt: []\newline Book: True
  & U: hello , i am looking for place -s to go in the [value\_area] .\newline S: there are many attractions in the [value\_area] like museums , architecture , boating , and @UNK . what are you interested in ?\newline U: how about a boating attraction ?\newline S: there are [value\_count] in the [value\_area] of town . [hotel\_name] , and [hotel\_name] . would either of those interest you ?\newline U: could you give me the address for [attraction\_name] , please ? i also need a place to stay , preferably somewhere [value\_pricerange] .\newline S: sure , [attraction\_name] is @UNK at [attraction\_address] . the [hotel\_name] is in the [value\_pricerange] price range with [value\_count] stars . would you like to book a room ?\newline U: does it offer free wifi internet ?
  & {\bf LM+G}: yes , it does . would you like me to book a room for you ? \newline{\bf HRED}: yes , it does have free parking . \newline {\bf {\ourmodel}}: yes , it does .
  \\
  \hline
\end{tabular}
\caption{Qualitative comparison of generated responses. Reqt: Requests. U: User, S: System.
  }
\label{table:respsample}
\end{table*}

